\documentclass[10pt,twocolumn,letterpaper]{article}

\usepackage[pagenumbers]{cvpr} 

\usepackage{graphicx}
\usepackage{amsmath}
\usepackage{amssymb}
\usepackage{booktabs}
\usepackage{url}

\usepackage[pagebackref,breaklinks,colorlinks]{hyperref}

\usepackage[capitalize]{cleveref}
\crefname{section}{Sec.}{Secs.}
\Crefname{section}{Section}{Sections}
\Crefname{table}{Table}{Tables}
\crefname{table}{Tab.}{Tabs.}

\def\cvprPaperID{*****} 
\def\confName{CVPR}
\def\confYear{2023}

\begin{document}

\title{A Boosted Model Ensembling Approach to Ball Action Spotting in Videos: The Runner-Up Solution to CVPR'23 SoccerNet Challenge}

\author{Luping Wang\thanks{Equal contributions} \qquad Hao Guo$^{\ast}$ \qquad Bin Liu\thanks{Corresponding author}\\
    Research Center for Applied Mathematics and Machine Intelligence\\
    Zhejiang Lab, Hangzhou 311121, China\\
    {\tt\small \{wangluping, guoh, liubin\}@zhejianglab.com}
}

\maketitle

\begin{abstract}
    This technical report presents our solution to Ball Action Spotting in videos. Our method reached second place in the CVPR'23 SoccerNet Challenge. Details of this challenge can be found at \url{https://www.soccer-net.org/tasks/ball-action-spotting}. Our approach is developed based on a baseline model termed E2E-Spot~\cite{e2e_spot}, which was provided by the organizer of this competition. We first generated several variants of the E2E-Spot model, resulting in a candidate model set. We then proposed a strategy for selecting appropriate model members from this set and assigning an appropriate weight to each model. The aim of this strategy is to boost the performance of the resulting model ensemble. Therefore, we call our approach Boosted Model Ensembling (BME). Our code is available at \url{https://github.com/ZJLAB-AMMI/E2E-Spot-MBS}.
\end{abstract}
\section{Introduction}
\label{sec:intro}
To better understand the salient actions of a broadcast soccer game, SoccerNet has introduced the task of action spotting, which involves finding all the actions occurring in the videos. This task addresses the more general problem of retrieving moments with specific semantic meaning in long untrimmed videos, extending beyond just soccer understanding. Details of the SoccerNet Ball Action Spotting challenge can be found at \url{https://www.soccer-net.org/tasks/ball-action-spotting}. In this technical report, we introduce our submitted solution termed Boosted Model Ensembling (BME), which reached second place in this Challenge..

Our proposed solution is built on a baseline model termed E2E-Spot~\cite{e2e_spot}, which was provided by organizers of this challenge. We analyzed E2E-Spot and identified 3 opportunities to improve it for addressing the SoccerNet Ball Action Spotting challenge:
\begin{enumerate}
    \item In the data set, only one frame associated with a representative event is labeled, whereas in reality, each considered event should be associated with multiple consecutive frames.
    \item A higher quality event feature extraction may help, as indicated by~\cite{feat_baidu}.
    \item the loss function used in training E2E-Spot is not totally consistent with the metric $\mbox{mAP@1}$, which is adopted by this challenge.
\end{enumerate}

Taking all above issues into consideration, we developed our solution BME, described in detail in Section~\ref{sec:method}. The experimental setting is presented in Section~\ref{sec:exp_set}, and some key results are showed in Section~\ref{sec:res}. Finally, we conclude our work in Section~\ref{sec:conclusion}.

\section{Our Method}
\label{sec:method}
In this section, we describe our proposed method BME in detail.

\subsection{The Model Ensembling Operation}
\begin{figure*}[htbp]
  \centering
   \includegraphics[width=\linewidth]{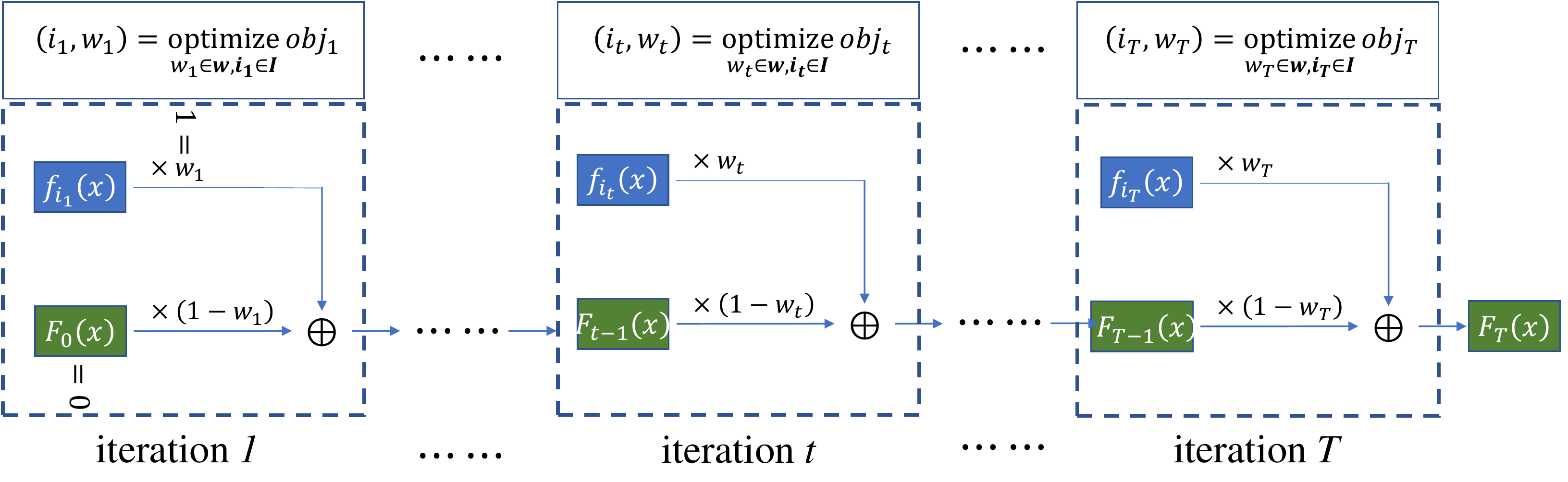}
   \caption{An Overview of the model ensembling operation in BME}
   \label{fig:overview}
\end{figure*}

The key operation of BME is model ensembling, which is illustrated in Figure~\ref{fig:overview}. As is shown, the final model ensemble $F_{T}$ is obtained after $T$ iterations. At an iteration say $t$, an objective function $obj_{t}$ associated with the performance metric ($\mbox{mAP@1}$ used here) is defined (namely, Equation (2) in Section 2.2), which is used to select the best model $f_{i_{t}}$ and its weight value $w_{t}$. Then, a new model ensemble $F_{t}$ is got by combining  $F_{t-1}$ and $f_{i_{t}}$ as follows

\begin{equation}\label{eq:ensemble}
    F_{t}(x) = (1 - w_{t})F_{t-1}(x) + w_{t}f_{i_{t}}(x)
\end{equation}

\subsection{Objective function}
The objective function used to select the best model $f_{i_{t}}$ and its weight value $w_{t}$, which appear in Equation (1), is defined as follows
\begin{equation}\label{eq:obj}
    obj_{t} = e(F_{t}, \mathcal{D}_{\text{valid}}) - e(F_{t-1}, \mathcal{D}_{\text{valid}})
\end{equation}
where the function $e(\cdot)$ denotes the target performance metric ($\mbox{mAP@1}$ here), $\mathcal{D}_{\text{valid}}$ the validation data set. At each iteration, we search for an appropriate member model and its corresponding weight that maximizes Equation (2), and then update the model ensemble according to Equation (1).

\subsection{Generating Candidate Models}
All candidate models are built on E2E-Spot~\cite{e2e_spot}. Their differences lie in: (1) the training samples being used; (2) network architectures for feature extraction; and (3) the optimizer being used.

\begin{figure*}[htbp]
  \centering
   \includegraphics[width=\linewidth]{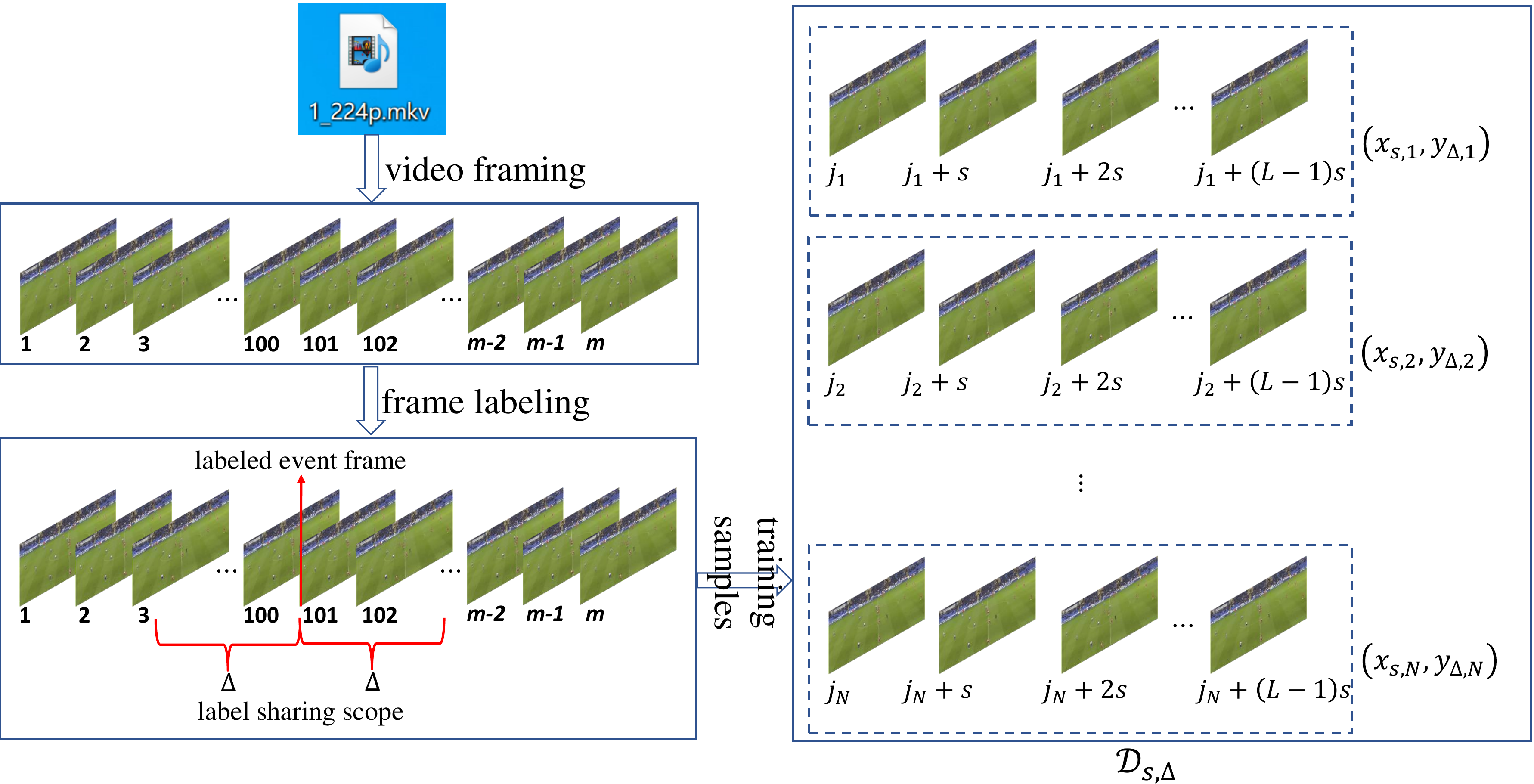}
   \caption{Our operation for generating different training data sets}
   \label{fig:train_sample}
\end{figure*}

\paragraph{Training samples}
We generate training samples as shown in Figure~\ref{fig:train_sample}. Firstly, the video is decomposed into a fixed number of frames per second ($FPS=25$ in our case). Then, all the achieved frames are labeled based on the time of the given events together with the label sharing scope controlled by a hyper-parameter $\Delta$. Finally, a training sample-set $\mathcal{D}_{s, \Delta}=\{(x_{s, i}, y_{\Delta, i})\}_{i=1}^{N}$ can be constructed by randomly picking $N$ video clips with a fixed clip length $L$ and a fixed frame stride size $s$. Different settings of $s$ and $\Delta$ lead to training sample sets with different properties. If $\Delta$ is set to a large value, the ratio of event frames increases while the ratio of error-labeled frames also increases and vice versa. The larger the stride size $s$, the longer time the clip covers, and the poorer continuity between the frames. Therefore, models trained with such different sample-sets will have different properties.

\paragraph{Network architectures for feature extraction}
The RegNet~\cite{regnet} is used as the baseline of the feature architecture in E2E-Spot~\cite{e2e_spot}. However, according to the experimental results reported in~\cite{e2e_spot}, RegNet performs worse than EfficientNet~\cite{enet} on the problems addressed in that paper. Therefore, RegNet and EfficientNet are the two candidates for the feature architecture considered in our solution. In addition, we incorporated the Gate-Shift Module (GSM)~\cite{gsm} into the 2D convolutional operator included in both RegNet and EfficientNet. The two versions of the feature architecture are denoted as $rny008\_gsm$ and $enetb2\_gsm$, respectively.

\paragraph{The optimizer}
We use the same baseline optimizer as in ~\cite{e2e_spot}, which is $\text{AdamW}$. In addition, we incorporate stochastic weight averaging (SWA)\cite{guo2023stochastic,swa} into the training process to improve the generalization of the trained model, denoted as $\text{AdamW}^{\dagger}$. Both $\text{AdamW}$ and $\text{AdamW}^{\dagger}$ are considered as candidates for the optimizer used to train a candidate model.

Each candidate model is trained with a specific combination of the training sample set, network architecture for feature extraction, and optimizer. Therefore, the number of candidate models is $N_{1} \times N_{2} \times N_{3}$, where $N_1, N_2,$ and $N_3$ represent the number of training sample sets, network architectures, and optimizers, respectively.

\section{Experimental setting}
\label{sec:exp_set}

\paragraph{Datasets} We solely used the dataset provided by the challenge organizers in our experiments. We employed five settings for constructing training samples, namely $\mathcal{D}_{1, 5}, \mathcal{D}_{1, 4}, \mathcal{D}_{2, 5}, \mathcal{D}_{2, 4}$, and $\mathcal{D}_{2, 2}$. The length of the clip is set to $L=100$, and the dimension of cropping for the frame is 224. During the test phase, we used $\mathcal{D}_{\text{valid}}$ as the validation dataset, while during the challenge phase, it was used as the test dataset.

\paragraph{Training candidate models}
All hyperparameters used for training candidate models were kept the same, unless otherwise specified. We selected GRU~\cite{gru} as the temporal architecture of E2E-Spot and employed data augmentation techniques such as random cropping, random flipping, brightness, contrast, hue, saturation, and MixUp~\cite{mixup} during training. The initial learning rate was set to 0.001 and was scheduled based on $LinearLR$ and $CosineAnnealingLR$ after warming up for 3 epochs. Each member model was trained for a total of 100 epochs on A100-GPU-80GB, with a batch size of 8. All the related source code was implemented using PyTorch 1.12.1.

\paragraph{Other issues}
During model inference, the length of overlap between adjacent clips is set to $L-1$, i.e., $overlap\_len=99$. After model inference, we employed non-maximum suppression (NMS)~\cite{nms} as a post-processing step on the predicted results. The window size, frame rate, and threshold of NMS are set to 10, 25, 0.01, respectively. When using BME to ensemble the sub-models, the weights are sampled from $\{0.1, 0.2, 0.3, \cdots, 1.0\}$.

\section{Results}
\label{sec:res}
To provide a clear view of the performance of each sub-model and the overall result of BME, we present the values of $\mbox{mAP@1}$ and the weights of the selected candidate models in Table~\ref{tab:sub_model}. From the table, we can observe that the performance of the sub-model candidates is similar, but their abilities and/or properties may differ. However, by combining the results of the selected sub-models through BME, we achieved a significant improvement, with an $\mbox{mAP@1}$ of \textbf{86.37\%}. These findings suggest that the method of generating candidate models is reasonable and the proposed BME approach is effective.

\begin{table*}[htbp]
  \centering
  \caption{The values of $\mbox{mAP@1}$ of all candidate models on the test dataset. The member models selected by BME is marked in bold and their corresponding weight values in the final model ensemble are marked in red.}
    \begin{tabular}{lllll}
    \toprule
          & \multicolumn{1}{p{5.735em}}{$rny008\_gsm$\newline{}$\text{AdamW}$} & \multicolumn{1}{p{5.735em}}{$rny008\_gsm$\newline{}$\text{AdamW}^{\dagger}$} & \multicolumn{1}{p{5.735em}}{$enetb2\_gsm$\newline{}$\text{AdamW}$} & \multicolumn{1}{p{5.735em}}{$enetb2\_gsm$\newline{}$\text{AdamW}^{\dagger}$} \\
    \midrule
    $\mathcal{D}_{1,5}$ & \textbf{82.82}(\textcolor{red}{0.081}) & 82.82 & \textbf{84.13}(\textcolor{red}{0.27}) & 83.86 \\
    $\mathcal{D}_{1,4}$ & \textbf{82.41}(\textcolor{red}{0.073}) & 82.62 & 83.7  & 83.46 \\
    $\mathcal{D}_{2,5}$ & 83.43 & \textbf{83.76}(\textcolor{red}{0.106}) & \textbf{83.49}(\textcolor{red}{0.066}) & 83.77 \\
    $\mathcal{D}_{2,4}$ & \textbf{83.61}(\textcolor{red}{0.09}) & \textbf{83.88}(\textcolor{red}{0.17}) & \textbf{83.28}(\textcolor{red}{0.059}) & \textbf{83.99}(\textcolor{red}{0.085}) \\
    $\mathcal{D}_{2,2}$ & 82.79 & 82.91 & 83.23 & 83.35 \\
    \bottomrule
    \end{tabular}%
  \label{tab:sub_model}%
\end{table*}%

\section{Conclusion}
\label{sec:conclusion}
In this report, we presented our submitted solution, termed Boosted Model Ensembling (BME), for the CVPR'23 SoccerNet Challenge (\url{https://www.soccer-net.org/tasks/ball-action-spotting}). BME is a model ensembling approach built on the end-to-end baseline model, E2E-Spot, as presented in~\cite{e2e_spot}. We generate several variants of the E2E-Spot model to create a candidate model set and propose a strategy for selecting appropriate model members from this set while assigning appropriate weights to each selected model. BME is characterized by operations for generating candidate models and a novel method for selecting and weighting them during the model ensembling process. The resulting ensemble model takes into account uncertainties in event length, optimal network architectures, and optimizers, making it more robust than the baseline model. Our approach can potentially be adapted to handle various video event analysis tasks.

{\small
\bibliographystyle{ieee_fullname}
\bibliography{egbib}
}

\end{document}